\documentclass{ecai} 

\usepackage{latexsym}
\usepackage{amssymb}
\usepackage{amsmath}
\usepackage{amsthm}
\usepackage{booktabs}
\usepackage{enumitem}
\usepackage{graphicx}
\usepackage{color}

\usepackage{multirow} 
\usepackage{makecell}
\usepackage{graphicx}  

\newcommand{\BibTeX}{B\kern-.05em{\sc i\kern-.025em b}\kern-.08em\TeX}

\begin{document}


\begin{frontmatter}


\paperid{5452}



\title{CheXLearner: Text-Guided Fine-Grained \\
Representation Learning for Progression Detection}


%

\author[A]{\fnms{Yuanzhuo}~\snm{Wang}}
\author[A]{\fnms{Junwen}~\snm{Duan}}
\author[A]{\fnms{Xinyu}~\snm{Li}}
\author[A]{\fnms{Jianxin}~\snm{Wang}\thanks{Corresponding Author. Email: jxwang@mail.csu.edu.cn.}}

\address[A]{School of Computer Science and Engineering, Central South University, Changsha, China}

\begin{abstract}
Temporal medical image analysis is essential for clinical decision-making, yet existing methods either align images and text at a coarse level—causing potential semantic mismatches—or depend solely on visual information, lacking medical semantic integration. We present \textbf{CheXLearner}, the first end-to-end framework that unifies anatomical region detection, Riemannian manifold-based structure alignment, and fine-grained regional semantic guidance. Our proposed Med-Manifold Alignment Module (Med-MAM) leverages hyperbolic geometry to robustly align anatomical structures and capture pathologically meaningful discrepancies across temporal chest X-rays. By introducing regional progression descriptions as supervision, CheXLearner achieves enhanced cross-modal representation learning and supports dynamic low-level feature optimization. Experiments show that CheXLearner achieves  81.12\% (+17.2\%) average accuracy and 80.32\% (+11.05\%) F1-score on anatomical region progression detection—substantially outperforming state-of-the-art baselines, especially in structurally complex regions. Additionally, our model attains a 91.52\% average AUC score in downstream disease classification, validating its superior feature representation. 
\end{abstract}

\end{frontmatter}

\section{Introduction}

Predicting disease progression in patients is essential for medical resource allocation, assisted diagnosis, and treatment planning. In recent years, with the continuous release of medical imaging datasets and advances in deep learning research, methods leveraging medical reports for assisted diagnosis have flourished. For example, \citep{liu2025enhanced} and \citep{liu2024structural} significantly improved image-text cross-modal representation quality through multi-view contrastive learning and knowledge-guided strategies. However, medical reports are inherently lists of localized findings rather than global descriptions. Existing methods (e.g., LaTiM \citep{zeghlache2024latim}, LOMIA-T \citep{sun2024lomia}) predominantly employ coarse-grained image-report level global contrastive learning, which, by forcing models to focus on global consistency, often overlooks local key feature associations (fine-grained connections between minor lesions and symptom phrases), thereby entailing risks of cross-modal mapping inaccuracies. \citep{li2024mlip,zhu2024multivariate}. Generative approaches (e.g., HistGen \citep{guo2024histgen}, KARGEN \citep{li2024kargen}) similarly produce hallucinated content (e.g., false-positive lesions) \citep{luo2024textual}, particularly evident in scenarios with limited historical data.

\begin{figure}[ht]
\centering
\includegraphics[width=1\linewidth]{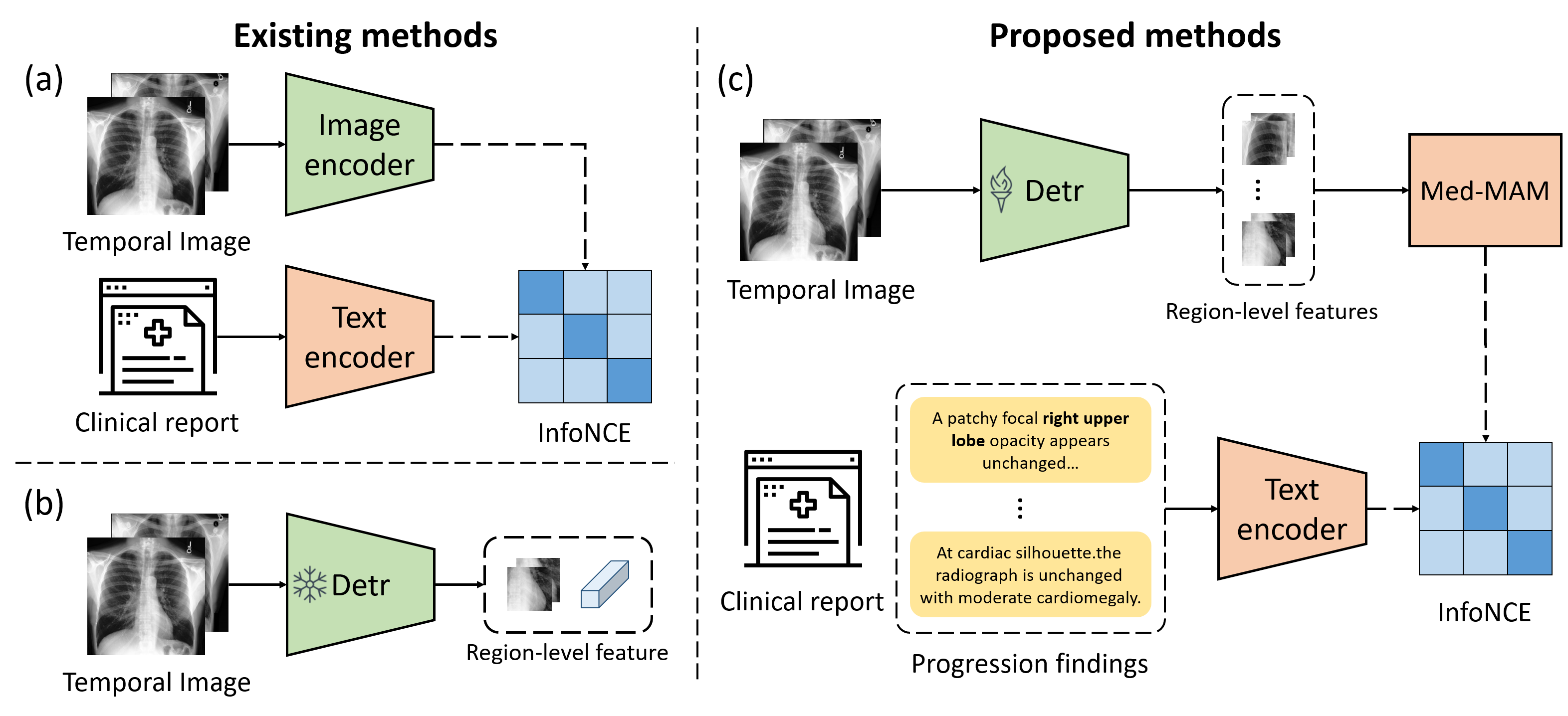}
\caption{(a) Methods (e.g., MLRG \citep{liu2025enhanced}, MLIP \citep{li2024mlip}) that rely on coarse image-report alignment. (b) Methods (e.g., CheXDetector) \citep{eshraghi2024representation} that  detect anatomical region progression but ignore semantic integration and suffer from regional noise. (c) Our proposed end-to-end model with anatomical structure alignment and regional progression semantics integration.}
\label{fig:model_compare}
\end{figure}

In contrast, region-based visual methods like CheXDetector \citep{eshraghi2024representation} achieve anatomical region-level disease progression analysis but rely solely on visual features. Without integration of medical text semantics, these methods are vulnerable to misregistration caused by imaging perspectives and physiological motion, particularly in fine-grained regions. Moreover, most current approaches use simplistic feature subtraction to extract differences, leading to loss of critical progression information. These limitations highlight three intertwined challenges: (1) the need for fine-grained alignment between textual descriptions of regional progression and the corresponding anatomical visual features; (2) the necessity for robust anatomical structure alignment to ensure accurate progression modeling; and (3) the representation bottleneck imposed by freezing parameters in two-stage frameworks.


To address these challenges, we propose CheXLearner, a structure-aligned, fine-grained, end-to-end multi-task framework for disease progression analysis. Our approach first explicitly aligns anatomical regions in sequential radiological images via DETR, then implicitly registers physiological structures using parallel transport on Riemannian hyperbolic manifolds. We further construct effective discrepancy features guided by region-level progression descriptions, enabling weakly supervised learning. The entire system is trained end-to-end, allowing comprehensive cross-modal optimization across both textual and progression classification tasks. As illustrated in Figure~\ref{fig:model_compare}, our framework fundamentally differs from prior works by integrating anatomical consistency modeling, region-level semantic supervision, and dynamic feature learning. Extensive experiments show CheXLearner achieves state-of-the-art performance in pulmonary disease progression detection (80.32\% F1-score, +11.05\%), with strong gains in structurally complex regions, and also yields highly representative features for downstream disease classification (91.52\% accuracy, +1.86\%). The main contributions of our work are summarized as follows:


\begin{itemize}
\item We propose CheXLearner, an end-to-end anatomical region-specific disease progression detection framework that optimizes visual feature learning through fine-grained cross-modal guidance signals derived from regional medical progression findings, achieving superior performance in regional disease progression detection tasks compared to baseline models.
\item Based on explicitly aligned anatomical regions, We design Med-MAM, a novel alignment module based on Riemannian hyperbolic manifold, which performs temporal anatomical structure alignment through parallel transport operations, effectively addressing domain discrepancies caused by imaging condition variations and physiological activities.
\item Under the joint constraints of anatomical structural alignment and fine-grained text-image alignment, we implement low-level feature optimization within an end-to-end framework and validate its effectiveness through experiments on downstream anatomical region disease classification tasks.
\end{itemize}

\section{Related Work}

\textbf{Temporal Medical Imaging} Early studies employed CNNs to process individual images \citep{singh2018deep,oh2019longitudinal} or RNNs (LSTMs/GRUs) for modeling regularly spaced longitudinal sequences \citep{santeramo2018longitudinal,xu2019deep}, but limited spatial feature modeling hindered fine-grained anatomical dynamics. With Transformer architectures, spatiotemporal fusion advanced through approaches like MoCo pre-training integrated with Transformers for multi-phase image features \citep{sriram2021covid}, graph attention networks modeling pathological co-occurrence statistics in CXRs (CheXRelNet) \citep{karwande2022chexrelnet}, and hierarchical multi-scale architectures (CheXRelFormer) \citep{mbakwe2023hierarchical}, though the latter relied on image registration and exhibited weak interpretability. Recent works incorporated textual data: BioViL-t \citep{bannur2023learning} aligned historical images with temporally relevant text phrases but suffered from 2D projections causing erroneous feature associations in 3D anatomy; Liu et al. (2025) enhanced report generation via multi-view contrastive learning but image-level modeling limited lesion localization \citep{liu2025enhanced}. Disease-specific modeling includes Chen et al. (2024)'s Alzheimer's progression analysis using MRI-derived deformation fields \citep{chen2024longformer}. Most relevant is CheXDetector \citep{eshraghi2024representation}, which localized 12 anatomical regions for progression tracking but lacked text integration and used a two-stage paradigm limiting joint feature modeling—gaps that motivate our work.

\textbf{Manifold Learning} Recent advances focus on cross-modal alignment in medical imaging. For cross-modal integration, MAGAN \cite{pmlr-v80-amodio18a} aligns single-cell RNA sequencing and mass cytometry data via adversarial training with correspondence loss, while DMDA \cite{yu2024bridging} mitigates global-local distribution discrepancies between whole-slide images (WSI) and mass spectrometry imaging (MSI) through geodesic flow kernel (GFK)-based feature migration. To address viewpoint/parameter interference, manifold correctors \cite{zeng2025medical} adjust image manifold distributions using spatial transformation networks, and geometry-preserving frameworks \cite{wang2013manifold} enforce multi-view consistency via cross-modal instance matching and global distance preservation. In semi-supervised segmentation, MANet \cite{shen2024manifold} improves boundary accuracy by embedding edge/surface features into manifolds, while M-CnT \cite{Huang_Huang_Xie_Lin_Tong_Chen_Li_Zheng_2024} aligns CNN-Transformer feature distributions using Frobenius norm metrics. Longitudinal analysis benefits from Riemannian hierarchical models \cite{9761465}, which implicitly align imaging trajectories to integrate spatiotemporal variations.

The mathematical foundations of manifold learning provide critical support for medical image analysis.\citep{pennec2006intrinsic} formalizes parallel transport mechanisms in tangent spaces for cross-manifold discrepancy vector migration, enabling intrinsic geometric analysis. \citep{konz2022intrinsic} identifies radiological images' heightened feature complexity and sensitivity to imaging viewpoints/device parameters compared to natural images, necessitating domain-specific manifold approaches. These works establish a geometric-to-clinical framework, forming the basis for our proposed viewpoint alignment methodology.These works forming the basis for our proposed anatomical structure alignment methodology.

\section{Method}

\begin{figure*}[ht]  
\centering
\includegraphics[width=\textwidth]{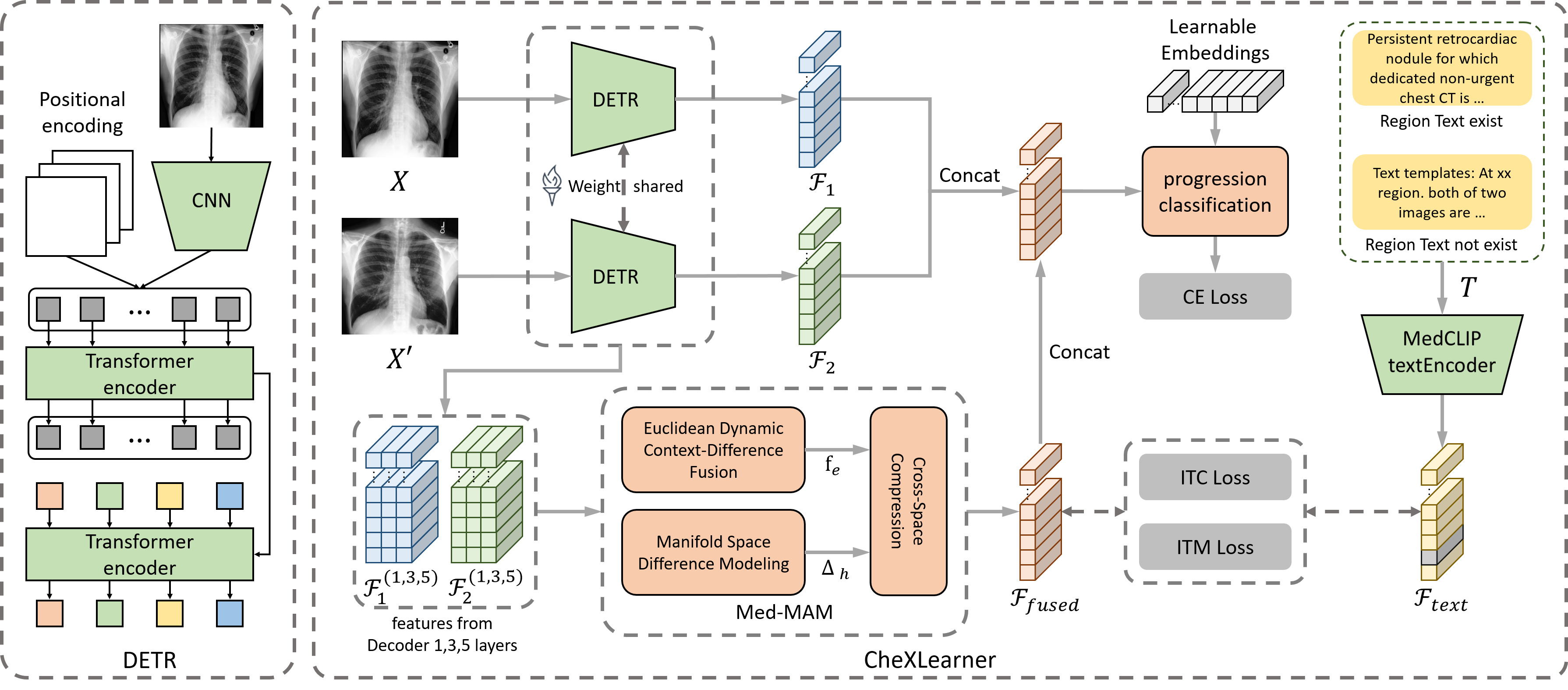}  
\caption{Overview of the CheXLearner framework. The model extracts region-of-interest (ROI) features from temporal CXRs using a DETR pretrained on CXR data. To address anatomical structure misalignment, the Med-MAM module mitigates domain shifts on Manifold and generates discrepancy features. Feature learning is facilitated through fine-grained image-text contrastive (ITC) and image-text matching (ITM) tasks that jointly model anatomical regions and disease progression description, with end-to-end optimization of the visual low-level features during training.}
\label{fig:model_pic}
\end{figure*}

CheXLearner is an end-to-end medical image analysis framework leveraging fine-grained textual guidance for disease progression detection, as illustrated in Figure~\ref{fig:model_pic}. The pipeline first localizes anatomical regions via DETR (Detection Transformer), then applies Med-MAM's hyperbolic geometric alignment (Section 3.3) to mitigate domain shifts, which generates discrepancy features. These features are optimized through cross-modal tasks (ITC/ITM) to learn semantic information of pathological variations, and finally aggregated by an attention module for final classification, with end-to-end optimization enhancing low-level feature representations.

\subsection{Problem Definition}

Let $C=\{(X, X')_i\}_{i=1}^{N}$ denote a collection of time-series CXR image pairs, where $X,X'\in \mathbb{R}^{H \times W \times C}$ represent two chest X-ray (CXR) images. Each image pair $(X, X')_i$ is associated with a regional progression label set $Y_i=\{y_{i,k}\}_{k=1}^K$ and corresponding text description set $T_i=\{t_{i,k}\}_{k=1}^K$. Here, $y_{i,k}$ indicates the disease progression label for the $k$-th anatomical region in image pair $(X,X')_i$, denoting whether the overall condition has improved, worsened, or no change. Meanwhile, $t_{i,k}$ represents the textual description of the $k$-th anatomical region for image pair $(X,X')_i$. The temporal medical image progression prediction task aims to predict the overall disease progression status across multiple anatomical regions between two sequential medical images.

\subsection{DETR Region Detection}

This study employs a pre-trained DETR \citep{carion2020end} model for anatomical region feature extraction. The model outputs a detection results set $B = \{(b_i, c_i, s_i)\}_{i=1}^K$ containing $K$ anatomical regions, where $b_i=(x_{\text{min}}, y_{\text{min}}, x_{\text{max}}, y_{\text{max}})$ denotes the bounding box coordinates of the $i$-th region, $c_i$ represents the predefined anatomical region category label, and $s_i$ indicates the region confidence score. Simultaneously, the last-layer hidden states $F = \{f_i\}_{i=1}^K$ of the model's decoder form anatomical region feature representations, where $f_i \in \mathbb{R}^{d}$ (with $d$ as the feature dimension) encodes information corresponding to the region.  

Unlike the two-stage decoupled training paradigm adopted by ChexDetector, we argue that optimizing underlying features for subsequent downstream tasks is critical. Specifically, we unfreeze all parameters of the pre-trained DETR model and achieve joint optimization from region detection to disease progression detection within an end-to-end training framework. This design enables the region detection module to dynamically adjust feature representations according to downstream task objectives.

\subsection{Med-Manifold Alignment Module(Med-MAM)}

After obtaining anatomical features, we introduce a Medical Manifold Alignment Module (Med-MAM) to address anatomical structure misalignment caused by detection errors, imaging variations, and physiological motion. The module is designed to integrate pathological discrepancy feature and contextual preservation in Euclidean space, align anatomical structures via a hyperbolic manifold with learnable curvature to mitigate cross-domain geometric deviations, and ultimately compress features across spaces to obtain effective fused representations.

\subsubsection{Euclidean Dynamic Context-Difference Fusion}

\begin{figure}[ht]
\centering
\includegraphics[width=7cm]{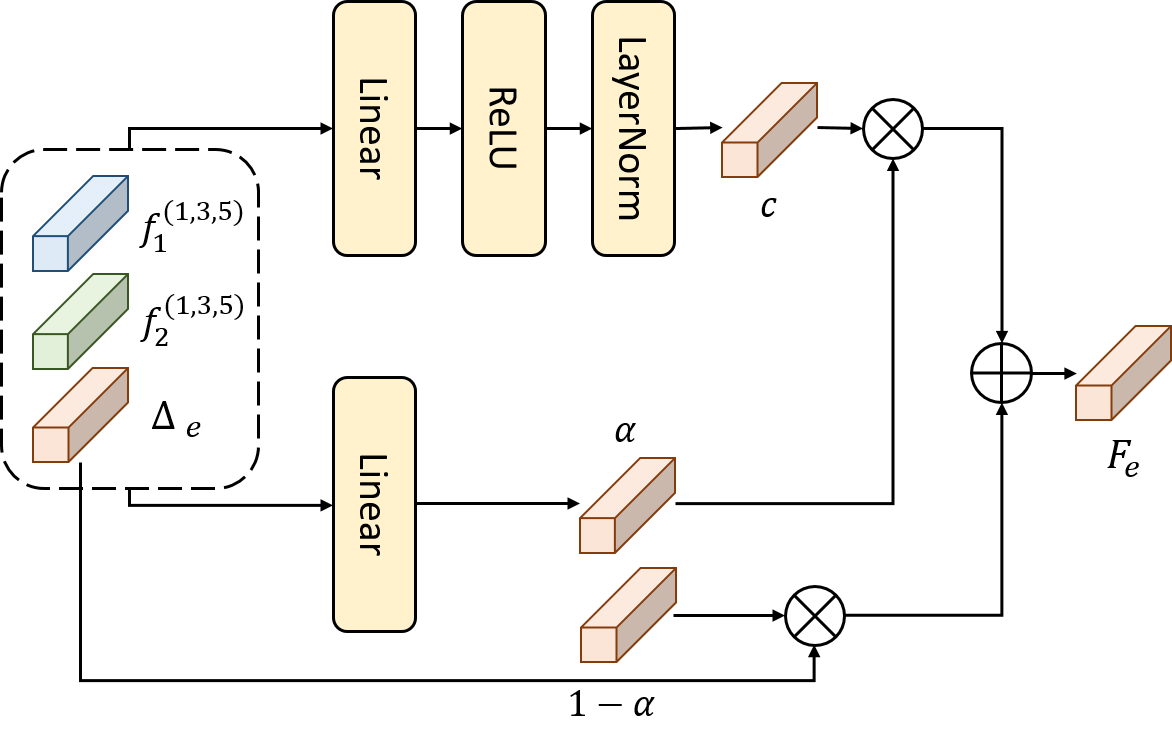}
\caption{Euclidean Dynamic Context-Difference Fusion module.}
\label{fig:fusion_module_part1}
\end{figure}

As illustrated in \ref{fig:fusion_module_part1}, this component constructs dynamic fusion of base features and difference features within Euclidean space. The input features $\mathbf{f}_1^{(1,3,5)}, \mathbf{f}_2^{(1,3,5)} \in \mathbb{R}^{3d}$ correspond to feature representations of the same anatomical region extracted from DETR's 1,3,5 decoder layers for image pairs $(X, X')_i$. The process begins with difference enhancement: the explicit difference vector $\Delta_e = \mathbf{f}_2^{(1,3,5)} - \mathbf{f}_1^{(1,3,5)} \in \mathbb{R}^{3d}$ is computed, followed by contrastive context construction:
\begin{eqnarray}\label{eq:eucli_context}
\mathbf{c} = \text{LN} \left( \text{ReLU} \left( \mathbf{W}_c \cdot \text{Concat} (\mathbf{f}_1^{(1,3,5)}, \mathbf{f}_2^{(1,3,5)}, \Delta_e \right) \right)
\end{eqnarray}

where $\mathbf{W}_c \in \mathbb{R}^{9d \times 3d}$ denotes a learnable weight matrix, $\text{Concat}$ represents concatenation, and $\text{LN}$ (Layer Normalization) normalizes along the feature dimension. Subsequently, an attention reweighting mechanism balances context preservation and difference enhancement. Learnable weight matrix $\mathbf{W}_a$ generate fusion weights:
\begin{eqnarray}\label{eq:eucli_atten_weight}
\alpha = \sigma \left( \mathbf{W}_a \cdot \text{Concat} (\mathbf{f}_1^{(1,3,5)}, \mathbf{f}_2^{(1,3,5)}, \Delta_e) + \mathbf{b}_1 \right)
\end{eqnarray}

where $\sigma$ is the Sigmoid function and $\mathbf{W}_a \in \mathbb{R}^{1 \times 9d}$ is a learnable parameter. The final fused feature $\mathbf{f}_e \in \mathbb{R}$ is obtained via weighted summation:
\begin{eqnarray}\label{eq:eucli_fused}
\mathbf{f}_e = \alpha \cdot \mathbf{c} + (1-\alpha) \cdot \Delta_e
\end{eqnarray}

This design dynamically coordinates contributions from global context and local differences through the adaptive weight $\alpha$.

\subsubsection{Manifold Space Difference Modeling}

\begin{figure}[ht]
\centering
\includegraphics[width=8.5cm]{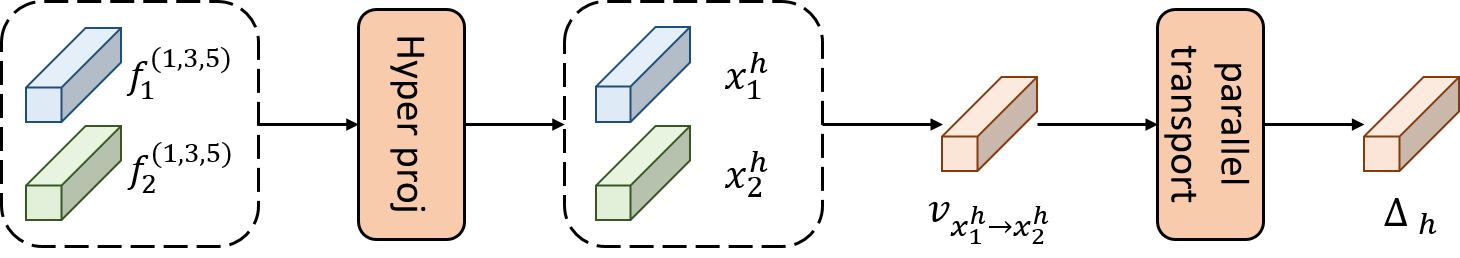}
\caption{Manifold difference modeling.}
\label{fig:fusion_module_part2}
\end{figure}

As illustrated in Figure~\ref{fig:fusion_module_part2}. Input features $\mathbf{f}_1^{(1,3,5)}$ and $\mathbf{f}_2^{(1,3,5)}$ are first mapped to the Poincaré manifold $\mathcal{M} = (\mathbb{D}_c, g)$ through dual-stream multilayer perceptrons:
\begin{eqnarray}\label{eq:manifold_proj}
\mathbf{x}_i^h = \Pi_{\mathcal{M}} \left( \text{MLP}(\mathbf{f}_i^{(1,3,5)}) \right), \quad i=1,2
\end{eqnarray}

where $\Pi_{\mathcal{M}}$ denotes manifold-constrained projection ensuring embedding into the hyperbolic space, with $\mathbb{D}_c$ representing the Poincaré disk model having learnable curvature parameter $c$. The projection operation is defined as:
\begin{eqnarray}\label{eq:manifold_proj_restraint}
\Pi_{\mathcal{M}}(\mathbf{z}) = \frac{\mathbf{z}}{\max(\|\mathbf{z}\|_2, 1/\sqrt{|c|})}
\end{eqnarray}

This projection embed anatomical hierarchies (e.g., organ-tissue-cell structures) into hyperbolic space with negative curvature via metric constraint $\|\mathbf{x}_i^h\|_2 < 1/\sqrt{c}$, leveraging its exponential volume growth to model hierarchical topological relationships inherent in medical images.

Disease progression features are extracted via Riemannian logarithm maps capturing geodesic differences between points on the manifold:
\begin{eqnarray}\label{eq:manifold_log_map}
v_{\mathbf{x}_1^h \to \mathbf{x}_2^h} = \log_{\mathbf{x}_1^h}(\mathbf{x}_2^h)
\end{eqnarray}

where $\log_{\mathbf{x}_1^h}(\cdot)$ maps point $\mathbf{x}_2^h$ to tangent space $T_{\mathbf{x}_1^h}\mathcal{M}$, with tangent vector $v$ geometrically representing geodesic direction/distance from $\mathbf{x}_1^h$ to $\mathbf{x}_2^h$. To mitigate domain shifts caused by imaging differences and physiological activities, we align discrepancy vectors through parallel transport to target domain tangent spaces:
\begin{eqnarray}\label{eq:manifold_diff}
\Delta_h = \Gamma_{\mathbf{x}_1^h \to \mathbf{x}_2^h}(v_{\mathbf{x}_1^h \to \mathbf{x}_2^h})
\end{eqnarray}

This operation converts discrepancy vectors from source domain perspective $\mathbf{x}_1^h$ to target domain perspective $\mathbf{x}_2^h$ along geodesics. Through parallel transport's conformal property (preserved angle between vector and geodesic) and length invariance ($\|\Gamma_{\mathbf{u} \to \mathbf{v}}(\mathbf{w})\| = \|\mathbf{w}\|$), we implicitly eliminate anatomical distortions while preserving pathological discrepancy. The parallel transport operator $\Gamma$ is defined as:
\begin{eqnarray}\label{eq:manifold_parallel_trans}
\Gamma_{\mathbf{u} \to \mathbf{v}}(\mathbf{w}) = \mathbf{w} - \frac{(1-\sqrt{c}\|\mathbf{v}\|)^2}{1 - c\langle\mathbf{u}, \mathbf{w}\rangle} \cdot \mathbf{v}
\end{eqnarray}

where $\mathbf{u}, \mathbf{v} \in \mathcal{M}$ are manifold points and $\mathbf{w} \in T_{\mathbf{u}}\mathcal{M}$ is the vector to transport. This preserves geometric consistency while implicitly eliminating cross-domain discrepancies.

%
%
%

\subsubsection{Cross-Space Compression}
The cross-space compression module processes two inputs: the Euclidean-space fused feature $\mathbf{f}_e \in \mathbb{R}^{3d}$ and manifold discrepancy feature $\Delta_h \in \mathbb{R}^{3d}$. These are concatenated into a $6d$-dimensional vector. A linear projection $\mathbf{W}_1 \in \mathbb{R}^{4d \times 6d}$ is then applied to reduce the dimensionality while preserving feature diversity:  
\begin{eqnarray}\label{eq:compression_layer1}
\mathbf{z}_1 = \text{LN}\left( \text{ReLU}\left( \mathbf{W}_1 \cdot \text{Concat}(\mathbf{f}_e, \Delta_h) + \mathbf{b}_2 \right) \right)
\end{eqnarray}

Subsequently, a second projection $\mathbf{W}_2 \in \mathbb{R}^{2d \times 4d}$ further compresses $\mathbf{z}_1$ to the target $2d$-dimension:  
\begin{eqnarray}
\label{eq:compression_layer2}
\mathbf{f}_{fused} = \mathbf{W}_2 \mathbf{z}_1 + \mathbf{b}_3
\end{eqnarray}

This two-stage design first extracts compact representations ($\mathbf{z}_1 \in \mathbb{R}^{4d}$) through nonlinear transformations and subsequently maps them to the desired output space.

\subsection{Text Encoder}
We employ a dynamic template strategy: leveraging DETR's anatomical classification outputs, standardized descriptors "At [region]" are automatically generated during input processing to distinguish features across regions with similar or the same descriptions. For healthy cases, we adopt "both of two images are healthy, there is no evident change" as the template.

To extract semantic representations from medical progression descriptions and guiding visual model learning, we employ the pre-trained MedCLIP \citep{wang2022medclip} text encoder. Input texts are tokenized with the Bio\_ClinicalBERT \citep{lee2020biobert} tokenizer (maximum length 128) and fed into the pre-trained MedCLIP encoder to generate contextualized embeddings. During encoding, hidden states from the 1st, 2nd, and final layers are aggregated using 1+2+Last pooling. A lightweight linear projection head subsequently reduces the feature dimension from 768 to $2d$ to align with the $2d$-dimensional fused visual features from the Med-MAM module (where $d$=256).  The final text features are denoted as $f_{text} \in \mathbb{R}^{2d}$

\subsection{Progression Detection}


We introduce a cross-attention mechanism for final progression classification, enhancing the model's perception capability of regional progression by leveraging DETR-predicted region labels to index learnable query vectors. After processing through two Transformer blocks, the features are fed into a fully connected layer classifier to accomplish the three-class classification task for disease progression.

\subsection{Training Objectives}

The model is jointly optimized through a multi-task learning framework for object detection, cross-modal alignment, and classification tasks. The total loss is composed as follows:  

The object detection component retains the standard loss of the DETR model (including classification Focal Loss, regression L1 and GIoU Loss) to ensure detection performance is not degraded by multi-task training. This loss is denoted as $\mathcal{L}_{\text{detr}}$.  

To leverage text features to guide visual feature learning, an InfoNCE \citep{oord2018representation} contrastive loss is designed to enhance semantic alignment between visual and text features. Given feature $\mathbf{f}_{fused, i} \in \mathbb{R}^{2d}$ of a pair of  Anatomical region generated via Med-MAM and progression description feature $\mathbf{f}_{text, j} \in \mathbb{R}^{2d}$ attained via text encoder, the normalized cosine similarity matrix is computed as:  
\begin{eqnarray}\label{eq:Loss_cos_similarity}
\mathbf{S}_{i,j} = \frac{\mathbf{f}_{fused, i} \cdot \mathbf{f}_{text, j}}{\|\mathbf{f}_{fused, i}\| \cdot \|\mathbf{f}_{text, j}\|}  
\end{eqnarray}

After scaling with a temperature parameter $\tau = 0.05$, the contrastive loss is formulated as:  
\begin{eqnarray}\label{eq:Loss_contrastive}
\mathcal{L}_{\text{contrast}} = -\frac{1}{N} \sum_{i=1}^N \log \frac{\exp(\mathbf{S}_{i,i}/\tau)}{\sum_{j=1}^N \exp(\mathbf{S}_{i,j}/\tau)}  
\end{eqnarray}

This loss enforces higher similarity for positive pairs (diagonal elements $i=j$) compared to negative pairs ($i \neq j$), thereby injecting textual semantics into visual feature learning. The temperature coefficient $\tau$ controls the model's focus on fine-grained details.  

For the anatomical disease progression detection task, a class-weighted cross-entropy loss is adopted:  
\begin{eqnarray}\label{eq:Loss_classification}
\mathcal{L}_{\text{cls}} = -\frac{1}{N} \sum_{i=1}^N \sum_{c=1}^C w_c \cdot y_{i,c} \log(p_{i,c})  
\end{eqnarray}

Here, $y_{i,c} \in \{0,1\}$ denotes ground-truth labels, $p_{i,c} \in [0,1]$ represents predicted probabilities, and inverse class frequency weights $w_c = \frac{N}{\sum_{i=1}^N y_{i,c}}$ are applied to mitigate class imbalance.  

The overall training objective combines all components:  
\begin{eqnarray}\label{eq:Loss_overall}
\mathcal{L}_{\text{total}} = \mathcal{L}_{\text{detr}} + \mathcal{L}_{\text{contrast}} + \mathcal{L}_{\text{cls}}
\end{eqnarray}

\section{Experiments}

\subsection{Datasets}
We use the publicly available the Chest-ImaGenome \citep{wu2021chest} dataset, which comprises two subsets: the gold-standard dataset validated by radiologists through manual annotation of 500 patient cases for benchmarking purposes, and the silver-standard dataset automatically generated via rule-based natural language processing (NLP) combined with CXR atlas-based bounding box detection techniques, covering 237,827 MIMIC-CXR \citep{johnson2019mimic} images. The annotation framework includes 29 anatomical locations (e.g., right lung, cardiac silhouette) and 1,256 attribute relationships (e.g., atelectasis, device presence), with approximately 670,000 localized contrast relationship annotations constructed through temporal sequence comparisons (e.g., "improved", "worsened").

\begin{table}[ht]
\caption{Dataset characteristics for progression labels with medical description per anatomical regions} 
\label{tab:dataset_characteristics}
\centering
\begin{tabular}{@{}l|ccccc@{}} 
\toprule
\makecell{Anatomical\\location}  
  & \makecell{improved} 
  & \makecell{no\\change} 
  & \makecell{worsened} 
  & \makecell{total}  \\ 
\midrule
cardiac silhouette       & 1242  & 6083  & 2275  & 9600   \\
left costophrenic angle  & 3387  & 3668  & 4568  & 11623  \\
left hilar structures    & 4919  & 3049  & 5184  & 13152  \\
left lower lung zone     & 4127  & 4378  & 6369  & 14874  \\
left mid lung zone       & 1520  & 1582  & 2459  & 5561   \\
left upper lung zone     & 458   & 586   & 698   & 1742   \\
cardiac silhouette       & 1242  & 6083  & 2275  & 9600   \\
right apical zone        & 500   & 1069  & 365   & 1934   \\
right costophrenic angle & 3416  & 4101  & 4503  & 12020  \\
right hilar structures   & 4951  & 3189  & 5228  & 13368  \\
right lower lung zone    & 3712  & 3937  & 5853  & 13502  \\
right mid lung zone      & 1430  & 1611  & 2491  & 5532   \\
right upper lung zone    & 598   & 850   & 943   & 2391   \\
\midrule
total                    & 30260 & 34103 & 40936 & 105299 \\ 
\bottomrule
\end{tabular}
\end{table}


For the anatomical region disease progression detection task, we adopted a data processing framework similar to \citep{eshraghi2024representation}, while introducing scene graph extraction to incorporate progression descriptions from medical reports as textual supplementation. Specifically, for each temporal CXR image pair comparison, we extracted target anatomical region progression labels and corresponding descriptions. Data partitioning followed the original protocol: the silver-standard dataset was split into 70\%/10\%/20\% training/validation/test sets, forming a progression dataset containing 9 target diseases, 35,646 CXR pairs, and 427,752 anatomical region pairs across 12 anatomical locations. The label distribution included three categories: "improvement", "worsened", and "no change". To ensure label quality, we excluded multi-labeled "conflicting" samples(4,080). For progression description texts, we obtained corresponding anatomical region descriptions from scene graphs of Chest-ImaGenome. For text-available samples, dataset statistics are shown in Table~\ref{tab:dataset_characteristics}. The "no progression description" samples (318,373) not presented in the table were treated as healthy "no change" cases, with standardized texts generated via dynamic template filling strategy (see Methods for details).

\subsection{Baselines}


We adopts CheXDetector \citep{eshraghi2024representation} as the primary baseline model. This model represents the only publicly available work specifically addressing anatomical region progression analysis in temporal medical imaging, serving as a critical benchmark for comparative evaluation. Additionally, we construct a DETR+MLP architecture as a supplementary comparison model. Furthermore, we incorporate two image-level progression detection models as secondary baselines: CheXRelNet \citep{karwande2022chexrelnet}, which leverages graph attention to model anatomical region dependencies, and CheXRelFormer \citep{mbakwe2023hierarchical}, a hierarchical Transformer-based approach for spatiotemporal feature extraction.

\subsection{Implementation Details}

We pre-trained the DETR model on a silver-standard dataset to improve adaptability to anatomical region detection and accelerate convergence. Training used the AdamW optimizer (batch size 16, weight decay $1 \times 10^{-4}$), with differential learning rates: $1 \times 10^{-5}$ for the backbone and $5 \times 10^{-5}$ for other modules, completed in 1 epoch.
For disease progression detection, we implemented the model in PyTorch Lightning with contrastive learning (InfoNCE loss, $\tau$=0.05) and manifold learning (curvature=0.1). The DETR backbone was fine-tuned at $1 \times 10^{-5}$, while other modules used $5 \times 10^{-5}$ learning rate and $1 \times 10^{-4}$ weight decay. A StepLR scheduler reduced learning rates to 30\% every 5 epochs, with batch size 4 over 20 epochs. All loss components were equally weighted ($\lambda$=1).
Downstream anatomical region disease detection followed CheXdetector's approach: ROI features from DETR were classified via a 7-layer MLP using Adam optimizer (batch size 256, learning rate $5 \times 10^{-4}$, weight decay $1 \times 10^{-5}$) over 15 epochs.
All experiments ran on a single NVIDIA RTX 3090 GPU using double-precision arithmetic for numerical stability, completing training pipelines in 36 hours.

\subsection{Experimental Results}

\subsubsection{Anatomical Location Detection Results}

Table~\ref{tab:detr_result} reports the AP (Average Precision) values for 12 anatomical regions under an IoU=0.5 threshold, with a mean Average Precision (mAP) of 91.36\% (CheXDetector baseline: 90.20\%), using the PASCAL VOC 2012 methodology \citep{everingham2015pascal}. The model achieves optimal performance in the upper lung zones: the right upper lung zone and left upper lung zone exhibit the highest AP values (98.00\% and 97.73\%), attributed to their relatively simple anatomical structures and clear spatial characteristics.

\begin{table}[ht]
\caption{DETR's detection performance across 12 anatomical regions.(Anatomical Location: Average Precision(\%), IoU Threshold=0.5). }
\label{tab:detr_result}
\centering
\begin{tabular}{@{}ll|ll@{}}
\toprule
anatomical Region          & AP(\%) & anatomical Region       & AP(\%) \\ \midrule
right upper lung zone      & 98.00  & left upper lung zone    & 97.73  \\
right mid lung zone        & 93.56  & left mid lung zone      & 94.05  \\
right lower lung zone      & 93.06  & left lower lung zone    & 90.97  \\
right costophrenic angle   & 81.35  & left costophrenic angle & 73.03  \\
right hilar structures     & 93.80  & left hilar structures   & 93.68  \\
right apical zone          & 95.86  & cardiac silhouette      & 91.20  \\ \bottomrule
\end{tabular}
\end{table}

A notable performance decline is observed in the costophrenic angle regions, with AP values of 73.03\% (left) and 81.35\% (right). This degradation may stem from rib occlusion, ambiguous boundaries with adjacent structures (e.g., diaphragm, cardiac silhouette), and the small spatial scale of these regions, which collectively hinder precise localization.All remaining anatomical regions maintain AP values above 90\%, demonstrating the model's robust localization capability for mainstream anatomical structures.

\subsubsection{Progression Detection Results}

\begin{table}[ht]
\caption{Model comparison for disease progression detetion}
\label{tab:result_image_level}
\centering
\begin{tabular}{ll|rr}
\toprule
model               & processing level & Acc            & F1-Score
\\ \midrule
CheXRelnet          & image level      & 46.80          & -        \\
CheXRelFormer       & image level      & 49.30          & -        \\
DETR+MLP            & region level     & 60.35          & 57.54    \\
CheXDetector        & region level     & 63.92          & 69.27    \\
CheXLearner(ours)   & region level     & \textbf{81.12} & \textbf{80.32}  
\\ \bottomrule
\end{tabular}
\end{table}

As shown in Table~\ref{tab:result_image_level}, our model achieved the best results among several current models for disease progression detection. Furthermore, we analyze the progression patterns across various anatomical regions in CXR, As shown in Table~\ref{tab:result}, our model (CheXLearner) achieved significant performance improvements in lesion progression detection across 12 anatomical regions. With weighted average accuracy (81.12\%) and F1-score (80.32\%) surpassing baseline methods DETR+MLP (60.35\%/57.54\%) and CheXDetector with regional attention mechanism (63.92\%/69.27\%), our approach demonstrates superior effectiveness. While maintaining stability in high-contrast regions (e.g., upper lung zone, right apical zone), we achieved breakthrough improvements in complex anatomical areas.

\begin{table}[ht]
\caption{Model performance comparison across anatomical locations (Accuracy/F1-Score)} 
\label{tab:result}
\centering
\begin{tabular}{@{}l|ccc@{}} 
\toprule
\makecell{Anatomical\\location}  
  & \makecell{DETR+MLP} 
  & \makecell{CheXDetector} 
  & \makecell{CheXLearner\\(Ours)} \\ 
\midrule
cardiac silhouette       & 64.57/71.89 & 77.29/79.93 & \textbf{83.47/83.85} \\
left costophrenic angle  & 37.73/43.38 & 46.35/53.11 & \textbf{77.97/78.15} \\
left hilar structures    & 22.97/14.10 & 32.59/33.30 & \textbf{71.10/69.53} \\
left lower lung zone     & 22.98/14.21 & 34.24/35.14 & \textbf{65.09/64.33} \\
left mid lung zone       & 65.50/71.41 & 76.42/77.88 & \textbf{85.47/83.36} \\
left upper lung zone     & 95.44/94.58 & 94.97/94.40 & \textbf{96.04/95.01} \\
right apical zone        & 97.03/95.88 & 96.86/95.80 & 96.03/95.64 \\
right costophrenic angle & 42.88/48.43 & 51.23/57.31 & \textbf{77.38/77.93} \\
right hilar structures   & 22.50/13.11 & 34.79/36.52 & \textbf{70.94/69.49} \\
right lower lung zone    & 24.00/18.70 & 36.41/39.27 & \textbf{67.47/66.74} \\
right mid lung zone      & 68.35/73.39 & 80.39/80.22 & \textbf{85.05/83.45} \\
right upper lung zone    & 94.56/93.21 & 95.31/93.54 & 94.45/93.64 \\
\midrule
Average                  & 60.35/57.54 & 63.92/69.27 & \textbf{81.12/80.32} \\

\bottomrule
\end{tabular}
\end{table}

For the physiologically complex left/right hilar structures characterized by dense vascular, bronchial, and lymphatic distributions causing lesion-normal tissue overlap, where baseline models performed worst, our Med-MAM module enabled effective anatomical alignment and pathological discrepancy capture, achieving maximum F1-Score improvements of +36.23\% and +32.97\%. Similarly, the left/right costophrenic angles, challenged by small annotation ranges and proximity to complex structures like diaphragm interfaces prone to imaging angle variations and detection errors, showed substantial gains of +25.04\% and +20.62\% through manifold space alignment.

The left/right lower lung zones are high-incidence regions for pneumonia exudation and pulmonary edema, where diffuse lesions frequently overlap with normal pulmonary textures or diaphragmatic shadows. Our model achieved F1-score improvements of 29.19\% and 27.47\% in these zones due to physiological structure alignment and fine-grained textual guidance. Despite these gains, performance remains suboptimal (64.33\%/66.74\%), indicating requirements for higher-resolution feature extraction.

Experimental results demonstrate that integrating anatomical structures alignment with fine-grained semantic alignment can effectively address the dual challenges of anatomical complexity and imaging perspective variations.Through an end-to-end joint optimization framework, this approach significantly enhances feature representation capabilities and disease progression detection performance.

\subsubsection{Downstream Disease Detection Results}

Table~\ref{tab:disease_detection} shows that our method  outperforms two-stage methods in downstream chest X-ray pathology classification tasks. Among the nine disease categories, inflammatory lesions with ambiguous boundaries such as Lung Opacity, Consolidation, and Pneumonia exhibit the most substantial performance improvements. The cross-modal alignment enhances semantic representation in non-specific regions through text-image interaction, while Riemannian manifold alignment improves feature robustness via parallel transport operations. Notably, anatomically distinct features like enlarged cardiac silhouette reach near-perfect AUC (0.99), suggesting performance saturation for anatomically defined pathologies. Compared to conventional two-stage methods, the proposed framework's integrated multi-modal learning avoids error cascades through end-to-end optimization, achieving 91.52\% average AUC (+1.86\%), demonstrating the effectiveness of multi-modal feature enhancement and geometric alignment strategies in medical image analysis.

\begin{table}[ht]
\caption{Results across 9 disease categories for the anatomical region disease detection task, a downstream application of DETR (AUC)}
\label{tab:disease_detection}
\centering
\begin{tabular}{@{}l|ccc@{}}
\toprule
disease                      & chexdetector & chexlearner & $\Delta$ \\ \midrule
Lung Opacity                 & 0.8390       & 0.8667      & 0.0277   \\
Pleural Effusion             & 0.9460       & 0.9550      & 0.0090   \\
Atelectasis                  & 0.9074       & 0.9143      & 0.0069   \\
Enlarged Cardiac Silhouette  & 0.9908       & 0.9912      & 0.0004   \\
Pulmonary Edema/Hazy Opacity & 0.8961       & 0.9226      & 0.0265   \\
Pneumothorax                 & 0.9145       & 0.9295      & 0.0150   \\
Consolidation                & 0.8611       & 0.8888      & 0.0277   \\
Fluid Overload/Heart Failure & 0.8747       & 0.8927      & 0.0180   \\
Pneumonia                    & 0.8403       & 0.8758      & 0.0355   \\ \midrule
average                      & 0.8966       & 0.9152      & 0.0186   \\ \bottomrule
\end{tabular}
\end{table}

\begin{figure*}[ht]  
\centering
\includegraphics[width=\textwidth]{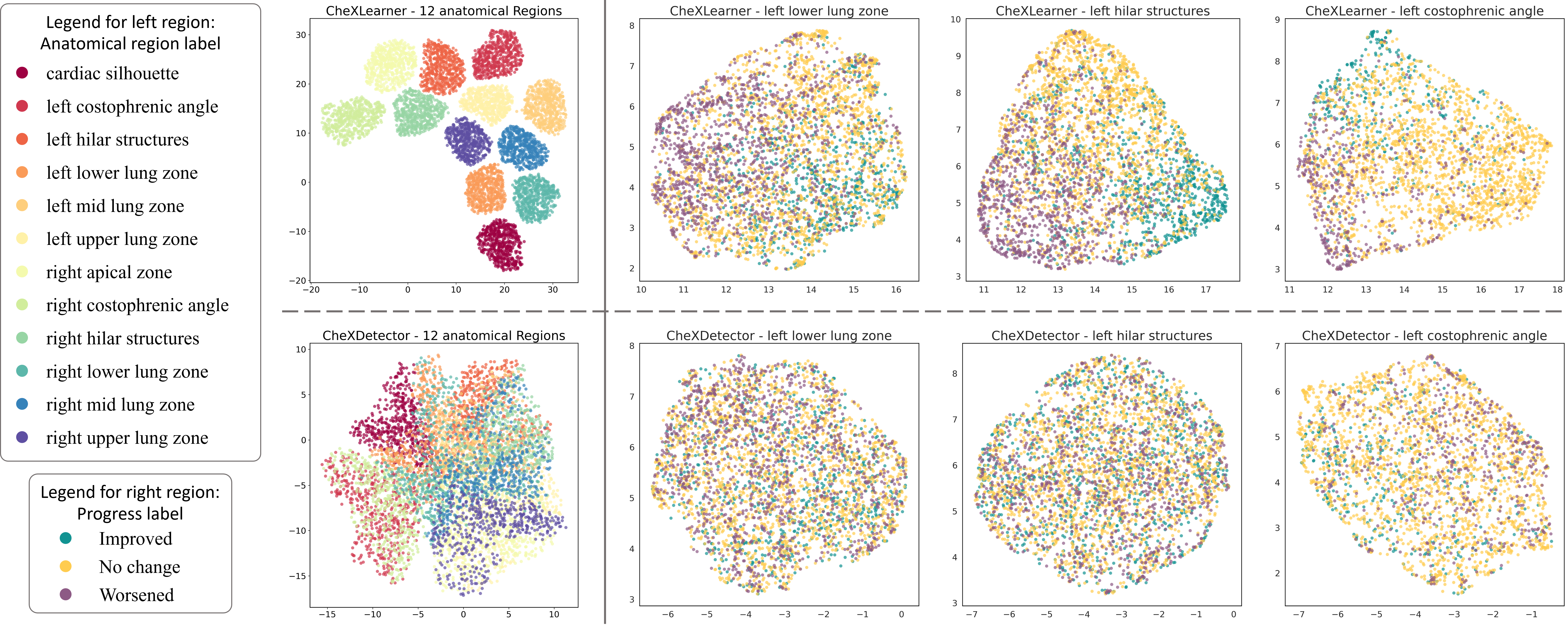}  
\caption{Scatter Plot of Feature Distributions for Our CheXLearner (upper region) vs. CheXDetector (lower region).}
\label{fig:scatter}
\end{figure*}

\subsection{Ablation Study}


\textbf{Ablation on Model Components} As shown in Table~\ref{tab:module_ablation}, unfreezing DETR parameters (Unfreeze\_DETR) delivers decisive performance gains. Using our base model as example, F1-Score improves by 7.22\% (66.59\%→73.81\%) after DETR optimization, demonstrating that adaptive low-level features are critical for cross-modal alignment. This pattern persists across architectures: even with ITC/Med-MAM modules, freezing DETR limits gains (3.24\% vs. 10.36\% improvement with unfreezing).

\begin{table}[ht]
\caption{Ablation study on model component (Acc/F1-Score)} 
\label{tab:module_ablation}
\centering
\begin{tabular}{@{}lc|cc@{}}
\toprule
model                         & unfreeze\_DETR & Acc   & F1-Score \\ \midrule
CheXDetector w/o region-atten & $\times$       & 60.18 & 66.74    \\
CheXDetector                  & $\times$       & 63.92 & 69.27    \\ \midrule
CheXLearner w/o ITC           & $\times$       & 60.22 & 66.59    \\
CheXLearner w/o ITC           & $\checkmark$   & 69.48 & 73.81    \\
CheXLearner w/ ITC            & $\times$       & 64.36 & 69.83    \\
CheXLearner w/ ITC            & $\checkmark$   & 80.84 & 80.19    \\
CheXLearner w/ ITM            & $\checkmark$   & 78.86 & 78.99    \\
CheXLearner w/ ITC\&ITM       & $\checkmark$   & 81.12 & 80.32    \\ \bottomrule
\end{tabular}
\end{table}


The Image-Text Contrastive (ITC) task achieves superior enhancement (+11.36\% Acc). Combined with Unfreeze\_DETR, medical text guidance boosts accuracy from 69.48\% to 80.84\%, validating that contrastive learning enables visual models to capture fine-grained discriminative features. Conversely, Image-Text Matching (ITM) yields slightly less 1.98\% gains (+9.38\% Acc), constrained by limited negative sample diversity and absence of hard negative mining. Joint ITC and ITM training provides marginal benefits (e.g., +0.28\% Acc), suggesting implicit redundancy in multi-task objectives.


\textbf{Ablation on Fusion Module} Table~\ref{tab:fusion_ablation} compares fusion strategies for progression detection. The feature difference method (x1-x2) performs the worst (66.25\% accuracy, 71.26\% F1-score), even underperforming the baseline model (69.48\% Acc, 73.81\% F1-Score). This reveals its failure to preserve semantic integrity and capture critical discrepancies. While feature concatenation (Concat) improves accuracy to 77.57\% by retaining more information, it lacks sufficient local interaction modeling.

\begin{table}[ht]
\caption{Ablation study on fusion modules for progression detection(\%)}
\label{tab:fusion_ablation}
\centering
\begin{tabular}{@{}ll|cc@{}}
\toprule
model & fusion\_module         & Acc     & F1-Score \\ 
\midrule
\multirow{5}{*}{CheXLearner}  
 & x1-x2                       & 66.25   & 71.26    \\
 & Concat                      & 77.57   & 78.48    \\
 & Transformer                 & 79.00   & 79.25    \\
 & Med-MAM w/o manifold        & 79.09   & 79.26    \\ 
 & Med-MAM                     & 81.12   & 80.32    \\ 
\bottomrule
\end{tabular}
\end{table}


Euclidean space struggles with handling rigid deformations (e.g., viewpoint variations) and elastic deformations (e.g., physiological activities). Approaches like Transformer (79.00\% Acc) and Med-MAM w/o manifold (79.09\% Acc) nearly reach their performance limits. Our proposed Med-MAM, by integrating contextual information and manifold space alignment, achieves 81.12\% Acc and 80.32\% F1, showing about 2\% Acc and 1\% F1 gains over Transformer. This demonstrates that manifold alignment effectively mitigate domain shifts and enhances cross-modal learning precision.

\subsection{Visual Analysis of Feature Representation}
We further extracted learned representations from both CheXLearner and CheXDetector on the test set and employed uniform UMAP parameter settings for dimensionality reduction to visualize the models' region-awareness (left region) and progression-awareness (right region), as illustrated in Figure~\ref{fig:scatter}. To prevent overcrowded visualizations, we applied balanced downsampling while preserving class distribution.

The results shows that features extracted by our model exhibit significant discriminability across various anatomical regions, with similar anatomical structures clustered closely together in the feature space, which can be attributed to the hyperbolic manifold framework integrated with cross-modal text alignment. In disease progression analysis, our model exhibits enhanced discrimination of samples across progression stages compared to baseline architectures. This suggests that anatomical structure alignment enables the emergence of pathologically meaningful features in the representation space.

\section{Conclusion}
In this paper, we propose an end-to-end CheXLearner model that effectively addresses anatomical structure misalignment issues after obtaining anatomical regions through object detection by innovatively introducing a Riemannian Manifold Alignment Module (Med-MAM). The model further employs fine-grained alignment of disease progression description to guide cross-modal alignment, significantly enhancing its capability in understanding regional disease progression. Experimental results demonstrate that CheXLearner exhibits significant advantages in regional disease progression analysis while optimizing the representation capability of underlying model features.Looking ahead, we will advance this research by integrating medical prior knowledge into a knowledge-enhanced module to address the current limitation of disease progression descriptions being conclusion-oriented with insufficient detail, and by extending global image analysis through multi-region integration and fine-grained alignment to broader medical imaging datasets.







\bibliography{mybibfile}

\end{document}